\documentclass{article}
\usepackage{spconf,amsmath,graphicx}
\usepackage{caption}
\usepackage{hyperref}

\title{ANOMALY LOCALITY IN VIDEO SURVEILLANCE}
\name{Federico Landi $^{1}$ \qquad
        Cees G. M. Snoek $^{2}$ \qquad
        Rita Cucchiara $^{1}$}
\address{$^{1}$ University of Modena and Reggio Emilia \qquad
        $^{2}$ University of Amsterdam}
        
\begin{document}
\ninept
\maketitle

\begin{abstract}
This paper strives for the detection of real-world anomalies such as \textit{burglaries} and \textit{assaults} in surveillance videos. Although anomalies are generally local, as they happen in a limited portion of the frame, none of the previous works on the subject has ever studied the contribution of locality. In this work, we explore the impact of considering spatiotemporal tubes instead of whole-frame video segments. For this purpose, we enrich existing surveillance videos with spatial and temporal annotations: it is the first dataset for anomaly detection with bounding box supervision in both its train and test set.
Our experiments show that a network trained with spatiotemporal tubes performs better than its analogous model trained with whole-frame videos. In addition, we discover that the locality is robust to different kinds of errors in the tube extraction phase at test time. Finally, we demonstrate that our network can provide spatiotemporal proposals for unseen surveillance videos leveraging only video-level labels. By doing, we enlarge our spatiotemporal anomaly dataset without the need for further human labeling.
\end{abstract}


\section{INTRODUCTION}
\label{sec:intro}
Anomalies are patterns that deviate from what is expected. When dealing with surveillance videos, anomalies consist of illegal behaviors or dangerous situations that could represent a threat to public safety. The need for automatic systems able to spot and report such events is strategic in real-world applications. For this reason, we aim to detect anomalies such as \textit{assaults}, \textit{robberies}, and \textit{burglaries} in videos. Recently, Sultani \textit{et al.} \cite{sultani2018real} propose a large dataset and a MIL-based solution for this challenging computer vision task, with the aim of bridging the gap between the recording capability of surveillance cameras and the limited number of human monitors. Where previous work on the topic only considered full-frame videos \cite{kim2009observe, mehran2009abnormal, liu2010anomaly, cui2011abnormal, zhao2011online, lu2013abnormal, xu2015learning, hasan2016learning, del2016discriminative, zhu2016anomaly, luo2017revisit}, we propose to exploit the inherent locality of anomalies and study whether the use of spatiotemporal tubes \cite{jain2014action} may help anomaly detection.

We draw inspiration from recent results on the importance of locality in other computer vision challenges, \textit{e.g.}, \cite{mettes2016spot, mettes2017localizing, cheron2018flexible}. Mettes \textit{et al.} show that even a single spatial point provides valuable information for action localization  \cite{mettes2016spot} and that it is possible to localize instances of actions leveraging only video-level labels and pseudo-annotations \cite{mettes2017localizing}. Ch\'eron \textit{et al.} \cite{cheron2018flexible} study the impact of different levels of supervision for action localization with a single, flexible model.
In their recent work, Hinami \textit{et al.} \cite{hinami2017joint} run a detector at each frame and compute the anomaly score related to visual concepts such as actions, objects, and attributes. They exploit an environment-specific model to identify unusual features that are likely to explain the abnormal pattern. Although their work employs locality, they only consider single frames instead of action tubes and do not study the benefit given by the detection. Like \cite{sultani2018real}, we also address anomaly detection as a regression problem and propose a model composed of a video encoder followed by a fully trainable regression network. Rather than relying on full-frame video segments, we integrate our model with a tube extraction module that lets the analysis focus on a particular set of spatiotemporal coordinates.

The first and chief contribution of this paper is the new approach to anomaly detection, based on action tubes instead of full frames. As a second contribution, we propose a new trainable model for anomaly detection designed to deal with different locations in the same video segment. Third, to show the potential of our approach, we enrich 100 surveillance videos from UCF-Crime \cite{sultani2018real} with spatiotemporal annotations for unusual events: UCFCrime2Local is the first dataset for anomaly detection with bounding box supervision in its train and test set. Finally, our experiments prove the importance of locality, the robustness of our model to different kinds of errors, and its reliability when adopted to provide weak annotations on new videos.
Before detailing our experiments, we discuss our model and the creation of UCFCrime2Local.

\section{LOCALITY IN ANOMALY DETECTION}
\label{sec:locality}
\begin{figure*}[!t]
    \includegraphics[width=\textwidth]{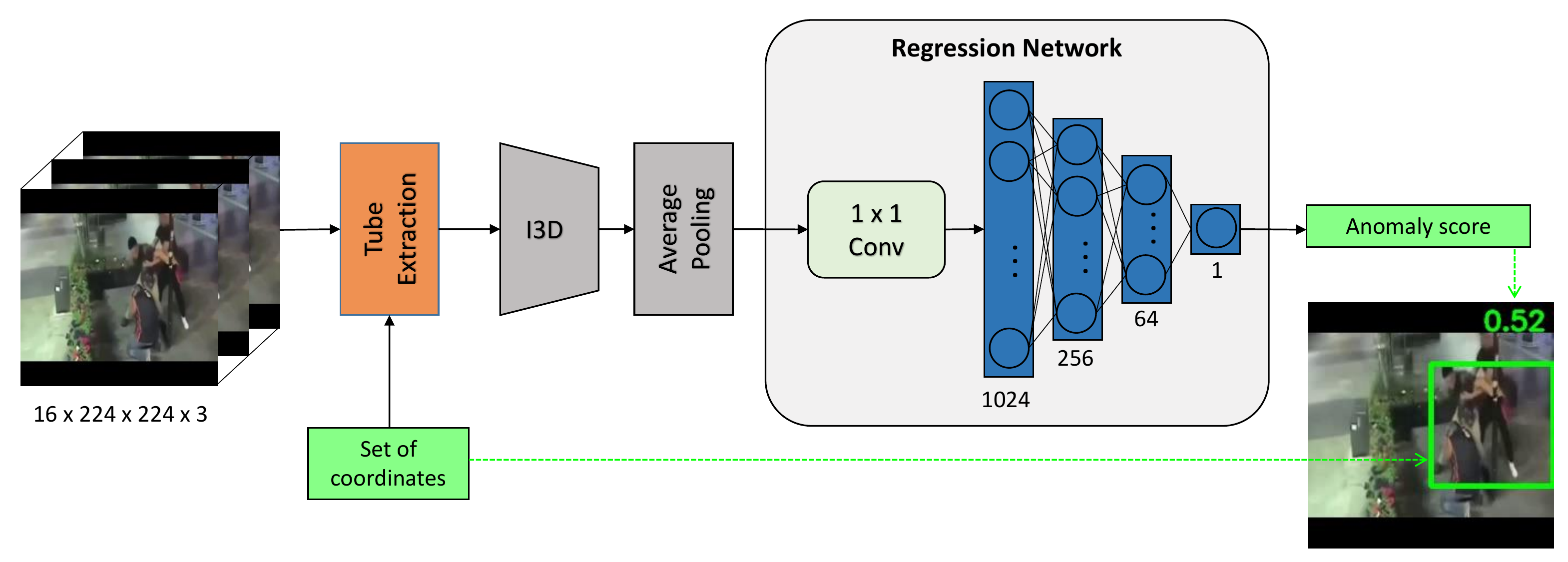}
    \caption{The flow diagram of the proposed method to embed locality in anomaly detection consists of three main components: a tube extraction module to consider a particular action tube instead of full frames, a 3D convolutional network \cite{carreira2017quo} to encode the video, and a regression network to map the input data into a single anomaly score.}
    \label{fig:fig1}
\end{figure*}

%The goal of this paper is to detect anomalies in surveillance videos. 
Given an input clip, we aim to determine whether the observed scene is normal or anomalous. Equivalently, we want our model to output the probability that an unusual event is taking place in the input video. The output of our model is a continuous number between 0 and 1, so we are casting anomaly detection to a regression problem, like \cite{sultani2018real}.
Additionally, we want to focus on the precise locality where the anomaly occurs: we do so by including a novel tube extraction module in our architecture. With this approach, we can change the granularity of our analysis from full-frame videos to spatiotemporal tubes. As depicted in Fig.\ \ref{fig:fig1}, our model consists of three main components: a tube extraction module, a video encoder, and a regression network.

\subsection{Tube Extraction}
\label{subsec:tube_extraction}
%We propose to study the impact of locality in anomaly detection. For this purpose, we need a mechanism to change the granularity of the investigation.
Given an input video and a set of coordinates, our tube extraction module produces a spatiotemporal volume as output.
%Different from \cite{hinami2017joint}, we do not run a detector to find bounding box coordinates, but we rely on hand-provided annotations from our UCFCrime2Local dataset, described in section \ref{sec:dataset}. In this way, we are not limited to single-person action tubes.
We implement tube extraction as a composition of crop and resize functions for each frame of the input video. Note that, if the coordinates match with the full frame, we do not focus on a precise location. We use this particular setup to evaluate our method on full-frame videos and provide a baseline in our experiments.

\textbf{Details.}
Given an input video composed of 16 frames with resolution $224\times224$ and RGB channels, thus having shape $16\times224\times224\times3$, we expect a set of 4 coordinates, identifying a bounding box, for each frame. We crop the portion of the video corresponding to the provided set of locations and then resize each frame to $224\times224$. Thus, input and output shapes of this block are equal.

\subsection{Video Encoder}
\label{subsec:video_encoder}
Given an input tube, we want to obtain a visual representation that encodes action information. Different from \cite{hinami2017joint, medel2016anomaly, liu2017future}, who use a 2D image-based architecture, we take advantage of the temporal dynamics of the video. Sultani \textit{et al.} \cite{sultani2018real} use C3D \cite{tran2015learning}, a 3D ConvNet, to extract features from a starting video segment. Recently, Carreira and Zisserman \cite{carreira2017quo} propose to inflate the 2D filters of a convolutional network to 3D kernels (I3D) to take advantage of the spatiotemporal nature of the video. By pre-training on ImageNet \cite{russakovsky2015imagenet} and Kinetics \cite{kay2017kinetics}, their model achieves state-of-the-art results for action recognition. We adopt I3D to encode information from our input video. We also follow the well-known two-stream approach \cite{simonyan2014two} to combine appearance and motion information, which was successfully applied earlier on other computer vision tasks such as action localization \cite{kalogeiton2017action}, and actor and action segmentation \cite{gavrilyuk2018actor}.
    
\textbf{Details.}
We extract features from the inception block before the last max-pooling layer, then we take an average pooling along the temporal dimension, similar to \cite{gavrilyuk2018actor}. The results of this operation is a feature cube with shape $14\times14\times832$. We apply the same procedure to optical flow, which is computed using the algorithm described by Farneb{\"a}ck \cite{farneback2003two}. We concatenate the two volumes, corresponding to RGB and flow, after the average pooling layer.  

\subsection{Regression Network}
\label{subsec:regression_network}
Given an input volume encoding the information about appearance and motion, our regression network outputs the anomaly score relative to the starting video. Since the score ranges between 0 and 1, we can interpret it as the probability that an unusual event is taking place in the investigated tube.
Given an input video $X$, its anomaly score $A(X)$ must comply with the following:
\begin{equation}
    \begin{cases} 
        0  \leq  A(X) < \boldsymbol{\tau} & \text{if }X\text{ is normal;} \\
        \boldsymbol{\tau} \leq A(X) \leq 1  & \text{if }X\text{ is anomalous.}
    \end{cases}
    \label{eq:cases}
\end{equation}
where the threshold $\boldsymbol{\tau}$ drives the binary classification into normal and anomalous videos.  Ideally, anomalous segments will score close to 1, while regular videos will map into values approaching 0. 

In our model, we perform regression with a convolutional layer, followed by a stack of fully-connected layers. The $1\times1$ convolution helps to preserve the dependencies over the feature maps along the spatial dimension while drastically reducing the number of parameters needed before the first fully-connected layer. 

\textbf{Details.}
We apply $1\times1$ convolution with 64 kernels. The output dimensions of the subsequent fully-connected layers are 1024, 256, 64, and 1. We use ReLU activation function and adopt 50\% dropout regularization \cite{srivastava2014dropout} between the FC layers.

\subsection{Training}
\label{subsec:training}
Our training sample consists of a 16-frame video segment $X$, a set of coordinates describing a spatiotemporal tube, denoted as $s$, and a ground truth label $y$. We employ the mean squared error as objective function:
\begin{equation}
    \mathcal{L} = \frac{1}{N} \sum _{i=1}^{N} {(A^s(X_i) - y_i)^2}
    \label{eq:mse}
\end{equation}
where $A^s(X_i) $ is the anomaly score relative to the action tube $s$ extracted from the input video $X_i$, and $y_i$ is a binary label.

\textbf{Details.}
We train our model using SGD optimizer with a learning rate of 0.001. We also adopt Nesterov momentum with intensity 0.9. We randomly select 5 video segments as a mini-batch, and we train for 10 epochs.  We compute the loss for each mini-batch as shown in Eq.\ \ref{eq:mse} and we back-propagate the gradients along the regression network.

\begin{figure*}[!t]
    \includegraphics[width=\textwidth]{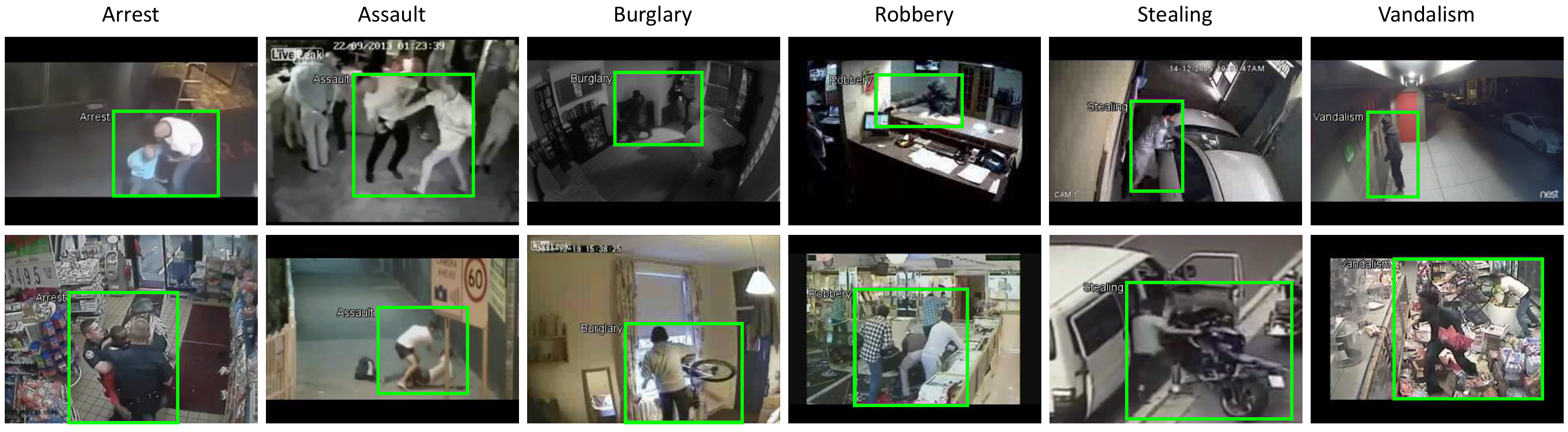}
    \caption{Examples of anomalies and ground-truth annotations in our dataset. In this paper, we do not use the anomaly category label.}
    \label{fig:fig2}
\end{figure*}

\section{OUR DATASET}
\label{sec:dataset}
To the best of our knowledge, none of the existing anomaly detection datasets \cite{sultani2018real, lu2013abnormal, datasetMinnesota, adam2008robust, mahadevan2010anomaly} provides spatiotemporal annotations for unusual events in its training set. To overcome the lack of labeled data, we enrich a portion of the recently-proposed UCF-Crime with spatiotemporal annotations. We start by selecting six among the 13 anomalous categories that are present in UCF-Crime, with particular attention to human-based anomalies: \textit{Arrest}, \textit{Assault}, \textit{Burglary}, \textit{Robbery}, \textit{Stealing}, and \textit{Vandalism}. We then select 100 videos belonging to the designated categories, resulting in more than an hour of video sequences. Finally, we annotate bounding boxes for anomalous events. 
%We are the first, to the best of our knowledge, to provide spatiotemporal annotations for this task. 
Although we do not use them in our experiments, action class labels for tasks such as action recognition or localization are available. After the annotation process, we add 200 negative clips from UCF-Crime. We leverage the fact that normal samples do not require further annotation. We split the dataset into training and testing set, respectively composed of 210 and 90 videos. Table \ref{table:dataset} reports some statistics about the videos in UCFCrime2Local. Our annotations are publicly available at \url{http://imagelab.ing.unimore.it/UCFCrime2Local}.

\subsection{Annotations}
\label{subsec:annotations}
Since the videos mainly contain human-based anomalies, a possibility to speed up the annotation would be to run a person detector for each frame and then merge the boxes along the temporal dimension. Although this approach is appealing, it suffers from many problems that make it inconvenient. First of all, it could not be generalized to annotate non-human events, such as \textit{road accidents} or \textit{explosions}. Second, not all the anomalies are single-person events: how could we capture the complexity of the interactions occurring in a fight scene? What about a robber threating a victim?
Moreover, while the availability of off-the-shelf detectors makes this choice appealing, a failure in the detection phase could seriously compromise the whole annotation process. For all of these reasons, we decide to annotate bounding boxes by hand. We choose Vatic \cite{vondrick2013efficiently} for his intuitive user interface, which can make the annotation relatively fast.

\textbf{Annotation Policy.}
We are aware that annotating spatiotemporal tubes is a delicate process that can be subject to many types of errors. For this reason, we adopt an annotation policy to enforce consistency in our work. An abnormal event comprehends the main characters directly performing the action and the secondary players that are eventually involved. For instance, one or more police officers are the main actors in an \textit{arrest} scene, while we include the captured person as a secondary actor. Concerning the temporal dimension, multiple anomalies of the same type can appear throughout an entire video: an anomaly begins when all the actors are visible, and it ends if they leave the scene. Regarding the spatial dimension, a single frame can contain only one bounding box, which should be large enough to include all the information about the anomalous action. For instance, in a \textit{stealing} situation, the stolen object should be included in the bounding box, as it provides relevant information about the action. Fig.\ \ref{fig:fig2} reports more examples from our annotations.

\begin{table}[!t]
    \centering
	\begin{tabular}{l|c|c}
	     & \textbf{Anomalous} & \textbf{Normal}   \\
	    \hline
		\textbf{Number of videos}       & 100 (69)  & 200 (141)  \\
		\textbf{Total length} (min)     & 66.3      & 112.1      \\
		\textbf{Average length} (sec)   & 39.8      & 33.6       \\
		\textbf{Min/Max length} (sec)  & 4.6/135.9 & 6.8/59.8
	\end{tabular}
	\caption{Statistics about UCFCrime2Local. Numbers in brackets represent the number of videos in the training set.}
	\label{table:dataset}
\end{table}

\section{EXPERIMENTS}
\label{sec:experiments}
\begin{figure*}[!t]
    \begin{minipage}[t]{0.29\linewidth}
        \centering
        \begin{tabular}{lc}
            \hline
             & \textbf{AUC (\%)} \\
            \textbf{Video Segment}      & 56.12             \\
            \textbf{Oracle Tube}        & \textbf{74.73}    \\
%            \textbf{Weakly-supervised}  & 80.96 \\
            \hline
        \end{tabular}
        \captionof{table}{Comparison between a full-\\frame and a tube-based approach.}
        \label{table:exp1}
    \end{minipage}
    \begin{minipage}[t]{0.71\linewidth}
        \centering
        \begin{tabular}{ccccccccccc}
            \hline
                & \multicolumn{2}{c}{\textbf{Smaller box}} & &
                \multicolumn{4}{c}{\textbf{Larger box}} &  & \multicolumn{2}{c}{\textbf{Translated box}} \\
                & 50\% & 75\% & &
                150\% & 200\% & 400\% & Full frame & &
                20px & 40px  \\
            \cline{2-3} \cline{5-8} \cline{10-11}
                \textbf{AUC (\%)} & 64.88 & 74.16 & &
                \textbf{77.52} & 77.37 & 74.38 & 68.20 &  &
                70.98 & 69.24
            \\ \hline
        \end{tabular}
        \captionof{table}{Robustness to localization errors in our locality-based approach. We show the AUC score for different settings. Our method is resilient against slightly larger boxes and modest cropping.}
        \label{table:exp2}
    \end{minipage}
\end{figure*}

%We aim to investigate the importance of locality in anomaly detection. In this set of experiments, we are the first, to the best of our knowledge, to explore the effects of using spatiotemporal tubes instead of full frames for this task. We validate our results on a new set of videos with bounding box annotation, explicitly provided for this purpose. 
%
%\textbf{Evaluation.}
%Like \cite{sultani2018real}, we make use of receiver operating characteristic (ROC) curve and the corresponding area under the curve (AUC) to evaluate our method. \vspace{-0.1cm}

\subsection{Full Frames Vs. Action Tubes}
\label{subsec:exp1}
In our first experiment, we test whether the use of locality can ease anomaly detection. Thanks to our flexible model and our annotations, we distinguish two settings:

%\noindent
\textbf{Video Segment.} We train our model on full frames without the information about the locality. In this setup, we can think of our tube extraction module as performing identity mapping from its input to the output. With a similar approach, we test this network on whole-frame videos.

%\noindent
\textbf{Oracle Tube.} During training, the coordinates given to the tube extractor is our ground-truth spatial annotation, hence the name Oracle Tube. We do the same during the test, challenging the network to discriminate between ordinary action tubes and atypical ones.

\textbf{Results.}
%\textbf{Evaluation.}
In Table \ref{table:exp1}, we report the results for this experiment. Like \cite{sultani2018real}, we use the receiver operating characteristic (ROC) curve and the corresponding area under the curve (AUC) to evaluate our method. The Oracle Tube setup, using locality, performs $18.61\%$ better than the Video Segment approach. This considerable improvement demonstrates that locality does ease anomaly detection.

\subsection{Robustness to Localization Errors}
\label{subsec:exp2}
In our second experiment, we progressively relax the ``oracle'' assumption during the testing phase of our Oracle Tube network. We aim to test the robustness of our model to localization errors, exploring at the same time the spatial extent of anomalies. We do that by perturbing the ground-truth coordinates supplied to our tube extraction module at test time with different kinds of errors. First, we reduce each side of the bounding boxes by a factor of $0.75$ $(75\%)$ and $0.50$ $(50\%)$. Second, we make the bounding boxes larger and larger until they incorporate the full frame. Finally, we use the original box dimension while translating the center from its ideal position. We apply translations of $20$ and $40$ pixels towards the top-left corner of the frame.

\textbf{Results.}
Numbers from Table \ref{table:exp2} show that a network trained with information about the locality performs better than the full-frame baseline even when dealing with localization errors at test time. In particular, we find out that for boxes dimensions in the range $75\%$ to $400\%$ than the ones we provided, results are very close to the ``ideal'' case, or even better. We conclude that adding some contextual information to action tubes helps anomaly detection.

\subsection{Independence and Reliability}
\label{subsec:exp3}
Finally, we want to test what occurs when we pick multiple action tubes from a video. Ideally, spatiotemporal volumes containing anomalies will score higher than regular ones. In this experiment, we let our model predict the anomaly scores for $N$ different tubes from each video segment in our testing set.
% We empirically set $N=61$.
In this way, we obtain $N$ anomaly scores for a single input clip, each of them related to a particular tube. We then apply different aggregation functions to determine the final score and report the results in Table \ref{table:exp3}.

\begin{figure}[!t]
    \centering
    \includegraphics[width=0.37\textwidth]{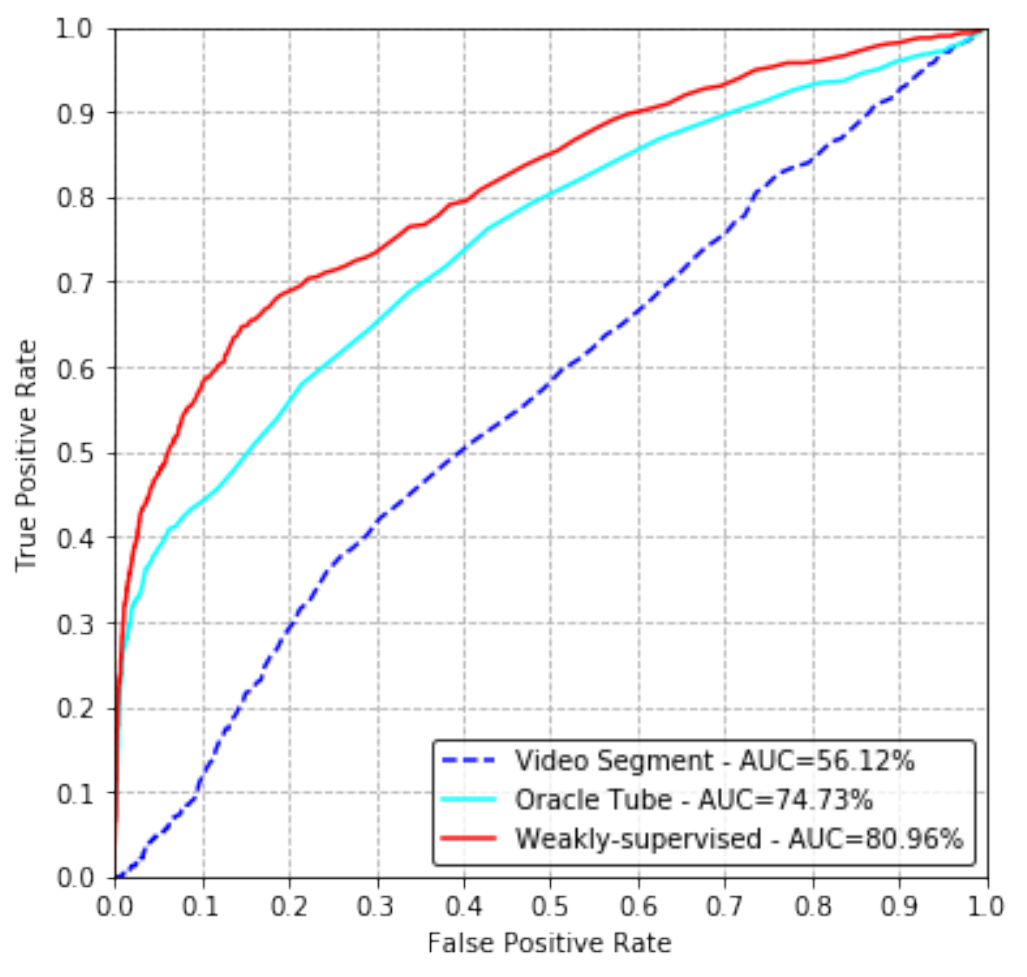}
    \caption{ROC comparison of full-frame approach (dashed, blue), Oracle Tube (cyan), and our weakly-supervised extension (red). We observe that: (1) we beat the full-frame baseline with a large margin, and (2) our weakly-supervised approach, having more training samples, outperforms Oracle Tube.}
    \label{fig:fig3}
\end{figure}

%\noindent
\textbf{Weakly-supervised Approach.}
The procedure described in the previous lines is a way to provide weak annotations for unseen videos, as it yields a set of spatiotemporal tubes and their corresponding anomaly scores. A simple method to test the reliability of our model is to use these proposals as weak annotations for training a weakly-supervised network. We collect for this purpose 100 new anomalous videos from UCF-Crime \cite{sultani2018real} belonging to the six categories described in Sec.\ \ref{sec:dataset}. We employ our model to get spatiotemporal proposals in the following way: we keep the highest score for each input clip along with its action tube, then we set the target label $y$ according to the rule:
\begin{equation}
    y =
    \begin{cases}
    0 & \text{if } A(X) < \boldsymbol{\tau_w}; \\
    1 & \text{if } A(X) \geq \boldsymbol{\tau_w}.
    \end{cases}
\end{equation}
where $\boldsymbol{\tau_w}$ is a threshold for our weakly-supervised proposals. In this experiment, we empirically set $\boldsymbol{\tau_w}=0.4$. We repeat the same process for negative samples, fixing the label $y=0$. We employ this new set of videos, coordinates, and labels to train our model from zero. In Fig.\ \ref{fig:fig3}, we compare our weakly-supervised extension to our previous settings trained with full supervision. 

\textbf{Results.}
Our weakly-supervised approach achieves better results than our strong-supervised network. This is not due solely to the proposals made by the latter, but also to the strong supervision that, though indirectly, influenced the training of our weakly-supervised extension. In this way, we show that proposals made over the new videos are valuable spatiotemporal annotations.

\begin{table}[!t]
    \resizebox{0.485\textwidth}{!} {
    \begin{tabular}{ccccc}
        \hline
        \textbf{Max} &
        \textbf{Top-10 avg} & \textbf{Top-20 avg} &
        \textbf{Top-30 avg} & \textbf{Top-45 avg} \\
        \hline
        66.94 &
        68.41 & \textbf{68.62} & 68.49 & 67.78 %& 64.19
        \\
        \hline
    \end{tabular} }
    \caption{AUC score for our tube-based approach when relaxing the ``oracle'' assumption. We score $N$ different tubes and then consider different aggregation functions for the corresponding scores. Our method performs $12.50\%$ better than the full-frame baseline when averaging the highest 20 scores for each video.}
    \label{table:exp3}
\end{table}

\section{CONCLUSION}
\label{sec:conclusion}
In this paper, we explore the importance of locality in anomaly detection. To that end, we enrich a portion of an existing dataset with spatiotemporal annotations, creating the first dataset for anomaly detection with bounding box supervision in both train and test set. We propose a new anomaly detection model for dealing with different spatiotemporal volumes in a single video. 
%Our method is the first, to the best of our knowledge, to exploit locality and action tubes for anomaly detection.
Experimental results show that: (1) locality helps anomaly detection; (2) our method is robust to different kinds of localization errors at test time and (3) it can provide spatiotemporal proposals over a potentially large set of unseen videos. These proposals are shown to be valuable for training a new weakly-supervised network. 
%Therefore, we conclude that our intuition about the inherent locality of real-world anomalies was right. 
We hope future work will follow our exploration, and consider locality when dealing with anomaly detection.

%%%%%%%%%%%%%%%%%%%  PUT REFERENCES ON 5TH PAGE ONLY  %%%%%%%%%%%%%%%%%
% BiBTeX files (here: strings, refs, manuals). The IEEEbib.bst 
% bibliography style file from IEEE produces unsorted bibliography list.
\bibliographystyle{IEEEbib}
\bibliography{references}

\end{document}